\documentclass[11pt,a4paper]{article}
\usepackage[hyperref]{acl2017}
\usepackage{times}
\usepackage{latexsym}
\usepackage{CJK, graphicx, multirow}

\usepackage{url}
\usepackage{amssymb}
\usepackage{balance}

\aclfinalcopy 
\def\aclpaperid{341} 

\title{Modeling Source Syntax for Neural Machine Translation}

\author{Junhui Li$^\dagger$\hspace{1cm}Deyi Xiong$^\dagger$\hspace{1cm}Zhaopeng Tu$^\ddagger$\thanks{~~Work done at Huawei Noah's Ark Lab, HongKong. }\\\textbf{Muhua Zhu}$^\ddagger$\hspace{1cm}\textbf{Min Zhang}$^\dagger$\hspace{1cm}\textbf{Guodong Zhou}$^\dagger$\\
        $^\dagger$School of Computer Science and Technology, \\Soochow University, Suzhou, China\\
        {\tt \{lijunhui, dyxiong, minzhang, gdzhou\}@suda.edu.cn}\\
        $^\ddagger$Tencent AI Lab, Shenzhen, China\\
        {\tt tuzhaopeng@gmail.com, muhuazhu@tencent.com}
       }

\date{}

\begin{document}
\begin{CJK}{UTF8}{gkai}
\maketitle
\begin{abstract}
Even though a linguistics-free sequence to sequence model in neural machine translation (NMT) has certain capability of implicitly learning syntactic information of source sentences, this paper shows that source syntax can be explicitly incorporated into NMT effectively to provide further improvements. Specifically, we linearize parse trees of source sentences to obtain structural label sequences. On the basis, we propose three different sorts of encoders to incorporate source syntax into NMT: 1) \textit{Parallel RNN} encoder that learns word and label annotation vectors parallelly; 2) \textit{Hierarchical RNN} encoder that learns word and label annotation vectors in a two-level hierarchy; and 3) \textit{Mixed RNN} encoder that stitchingly learns word and label annotation vectors over sequences where words and labels are mixed. Experimentation on Chinese-to-English translation demonstrates that all the three proposed syntactic encoders are able to improve translation accuracy. It is interesting to note that the simplest RNN encoder, i.e., \textit{Mixed RNN} encoder yields the best performance with an significant improvement of 1.4 BLEU points. Moreover, an in-depth analysis from several perspectives is provided to reveal how source syntax benefits NMT. 
\end{abstract}

\section{Introduction}\label{sect:intr}
Recently the sequence to sequence model (seq2seq) in neural machine translation (NMT) has achieved certain success over the state-of-the-art of statistical machine translation (SMT) on various language pairs~\cite{bahdanau_etal:15,jean_etal:15,luong_etal_emnlp:15,luong_etal_iwslt:15}. However, ~\citeauthor{shi_etal:16}~\shortcite{shi_etal:16} show that the seq2seq model still fails to capture a lot of deep structural details, even though it is capable of learning certain implicit source syntax from sentence-aligned parallel corpus. Moreover, it requires an additional parsing-task-specific training mechanism to recover the hidden syntax in NMT. As a result, in the absence of explicit linguistic knowledge, the seq2seq model in NMT tends to produce translations that fail to well respect syntax. In this paper, we show that syntax can be well exploited in NMT explicitly by taking advantage of source-side syntax to improve the translation accuracy.

\begin{figure}
\begin{center}
\includegraphics[width=3.0in]{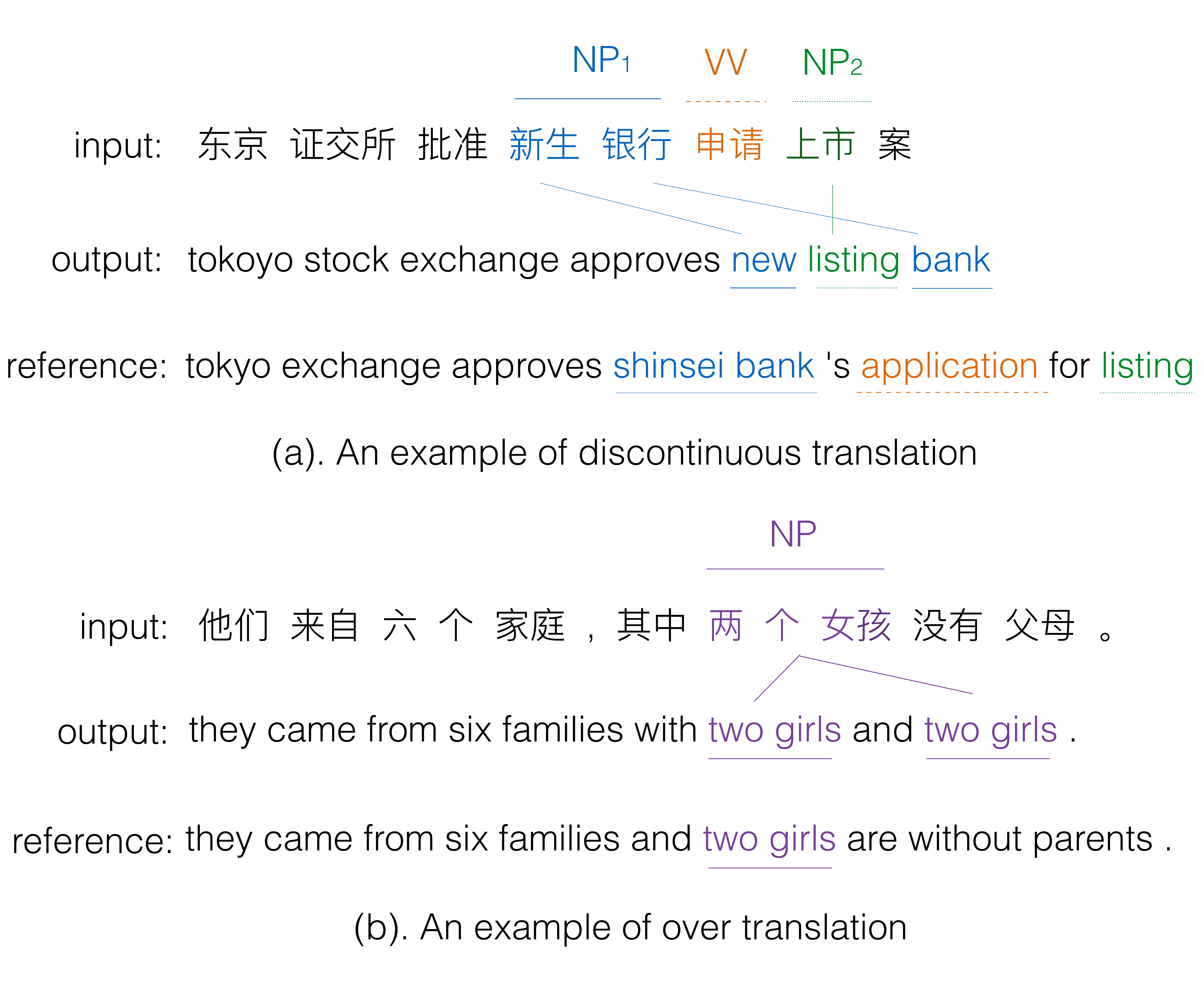}
\end{center}
\caption{Examples of NMT translation that fail to respect source syntax.}\label{fig:example}
\end{figure}

In principle, syntax is a promising avenue for translation modeling. This has been verified by tremendous encouraging studies on syntax-based SMT that substantially improves translation by integrating various kinds of syntactic knowledge~\cite{liu_etal:06,marton_resnik:08,shen_etal:08,li_etal:13}. While it is yet to be seen how syntax can benefit NMT effectively, we find that translations of NMT sometimes fail to well respect source syntax. Figure~\ref{fig:example} (a) shows a Chinese-to-English translation example of NMT. In this example, the NMT seq2seq model incorrectly translates the Chinese noun phrase (i.e., \textit{新生/xinsheng 银行/yinhang}) into a discontinuous phrase in English (i.e., \textit{new ... bank}) due to the failure of capturing the internal syntactic structure in the input Chinese sentence. Statistics on our development set show that one forth of Chinese noun phrases are translated into discontinuous phrases in English, indicating the substantial disrespect of syntax in NMT translation.\footnote{Manually examining 200 random such discontinuously translated noun phrases, we find that 90\% of them should be continuously translated according to the reference translation.} Figure~\ref{fig:example} (b) shows another example with over translation, where the noun phrase \textit{两/liang 个/ge 女孩/nvhai} is translated twice in English. Similar to discontinuous translation, over translation usually happens along with the disrespect of syntax which results in the repeated translation of the same source words in multiple positions of the target sentence.  

In this paper we are not aiming at solving any particular issue, either the discontinuous translation or the over translation. Alternatively, we address how to incorporate explicitly the source syntax to improve the NMT translation accuracy with the expectation of alleviating the issues above in general. Specifically, rather than directly assigning each source word with manually designed syntactic labels, as~\citeauthor{sennrich_haddow:16}~\shortcite{sennrich_haddow:16} do, we linearize a phrase parse tree into a \textit{structural label sequence} and let the model automatically learn useful syntactic information. On the basis, we systematically propose and compare several different approaches to incorporating the label sequence into the seq2seq NMT model. Experimentation on Chinese-to-English translation demonstrates that all proposed approaches are able to improve the translation accuracy.

\section{Attention-based NMT}\label{sect:nmt}
\begin{figure}
\setlength{\abovecaptionskip}{0pt}
\begin{center}
\includegraphics[width=3.0in]{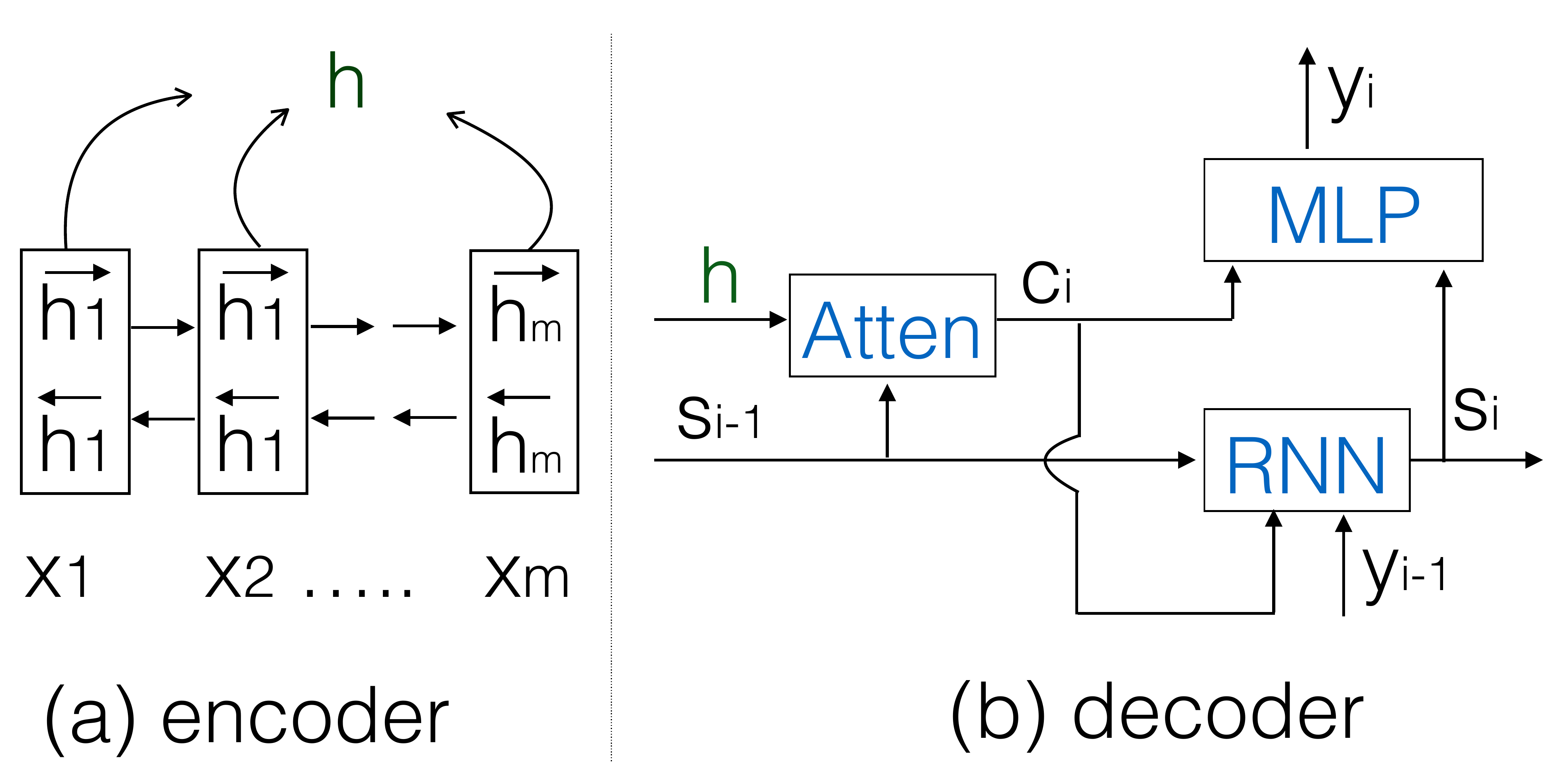}
\end{center}
\caption{Attention-based NMT model.}\label{fig:nmt}
\end{figure}

As a background and a baseline, in this section, we briefly describe the NMT model with an attention mechanism by ~\citeauthor{bahdanau_etal:15}~\shortcite{bahdanau_etal:15}, which mainly consists of an encoder and a decoder, as shown in Figure~\ref{fig:nmt}.

\vspace{0.2cm}
\noindent\textbf{Encoder} The encoding of a source sentence is formulated using a pair of neural networks, i.e., two recurrent neural networks (denoted \textit{bi-RNN}): one reads an input sequence $x = (x_1, ..., x_m)$ from left to right and outputs a forward sequence of hidden states $(\overrightarrow{h_1}, ..., \overrightarrow{h_m})$,  while the other operates from right to left and outputs a backward sequence $(\overleftarrow{h_1}, ..., \overleftarrow{h_m})$. Each source word $x_j$ is represented as $h_j$ (also referred to as word annotation vector): the concatenation of hidden states $\overrightarrow{h_j}$ and $\overleftarrow{h_j}$. Such bi-RNN encodes not only the word itself but also its left and right context, which can provide important evidence for its translation. 

\vspace{0.2cm}
\noindent\textbf{Decoder} The decoder is also an RNN that predicts a target sequence $y=(y_1, ..., y_n)$. Each target word $y_i$ is predicted via a multi-layer perceptron (MLP) component which is based on a recurrent hidden state $s_i$, the previous predicted word $y_{i-1}$, and a source-side context vector $c_i$. Here, $c_i$ is calculated as a weighted sum over source annotation vectors $(h_1, ..., h_m)$. The weight vector $\alpha_{i} \in\mathbb{R}^{m}$ over source annotation vectors is obtained by an attention model, which captures the correspondences between the source and the target languages. The attention weight $\alpha_{ij}$ is computed based on the previous recurrent hidden state $s_{i-1}$ and source annotation vector $h_j$.

\section{NMT with Source Syntax}\label{sect:syntax}
The conventional NMT models treat a sentence as a sequence of words and ignore external knowledge, failing to effectively capture various kinds of inherent structure of the sentence. To leverage external knowledge, specifically the syntax in the source side, we focus on the parse tree of a sentence and propose three different NMT models that explicitly consider the syntactic structure into encoding. Our purpose is to inform the NMT model the structural context of each word in its corresponding parse tree with the goal that the learned annotation vectors $(h_1, ..., h_m)$ encode not only the information of words and their surroundings, but also structural context in the parse tree. In the rest of this section, we use English sentences as examples to explain our methods.

\subsection{Syntax Representation}

\begin{figure}
\setlength{\abovecaptionskip}{0pt}
\begin{center}
\includegraphics[width=2.0in]{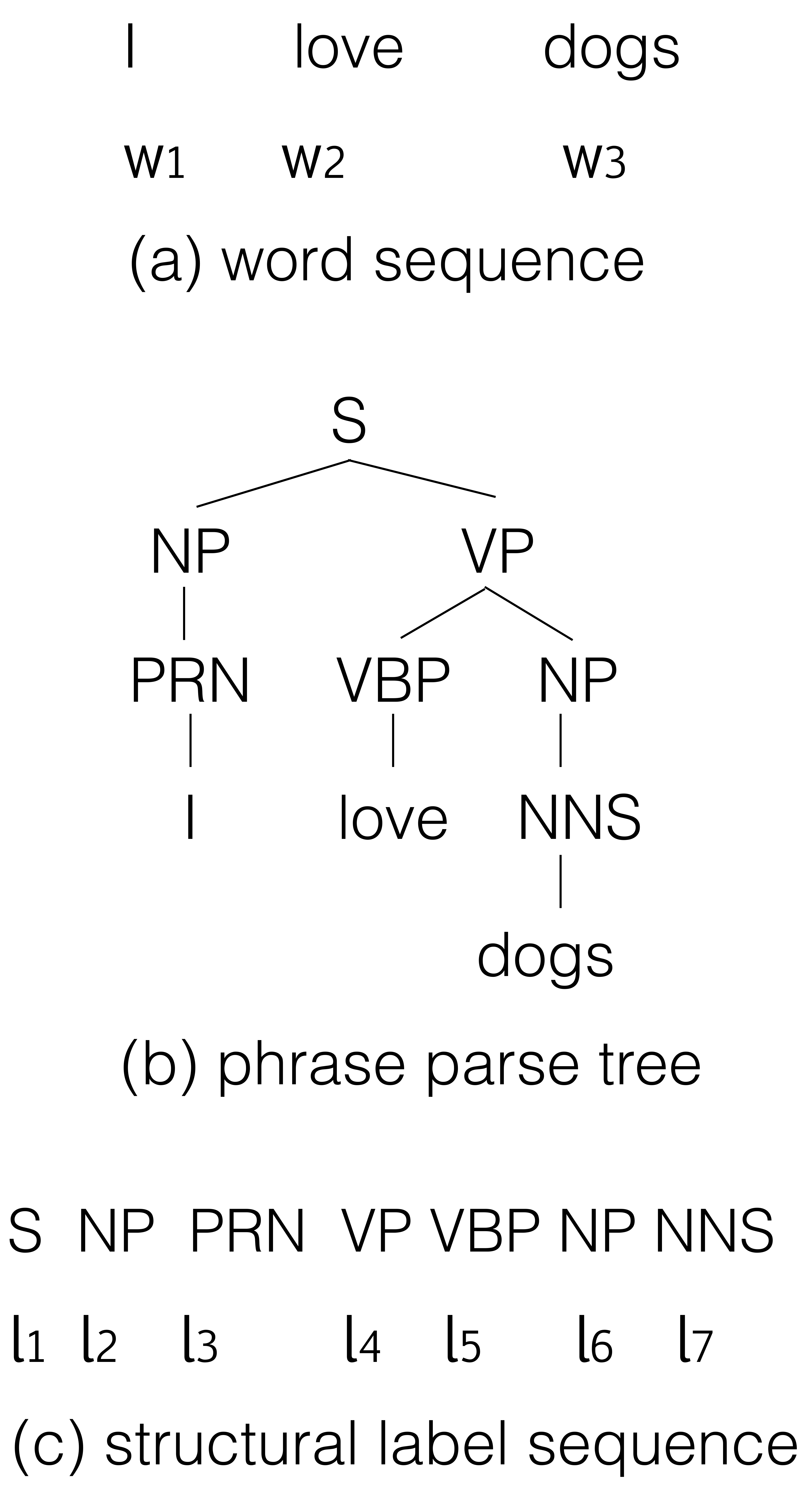}
\end{center}
\caption{An example of an input sentence (a), its parse tree (b), and the parse tree's sequential form (c).}\label{fig:parse_tree}
\end{figure}

To obtain the structural context of a word in its parse tree, ideally the model should not only capture and remember the whole parse tree structure, but also discriminate the contexts of any two different words. However, considering the lack of efficient way to directly model structural information, an alternative way is to linearize the phrase parse tree into a sequence of structural labels and learn the structural context through the sequence. For example,
Figure~\ref{fig:parse_tree}(c) shows the structural label sequence of Figure~\ref{fig:parse_tree}(b) in a simple way following a depth-first traversal order. Note that linearizing a parse tree in a depth-first traversal order into a sequence of structural labels has also been widely adopted in recent advances in neural syntactic parsing~\cite{vinyals_etal:15,choe_charniak:16}, suggesting that the linearized sequence can be viewed as an alternative to its tree structure.\footnote{We have also tried to include the ending brackets in the structural label sequence, as what~\cite{vinyals_etal:15,choe_charniak:16} do. However, the performance gap is very small by adding the ending brackets or not.} 

\begin{figure}[!ht]
\setlength{\abovecaptionskip}{0pt}
\begin{center}
\includegraphics[width=3.0in]{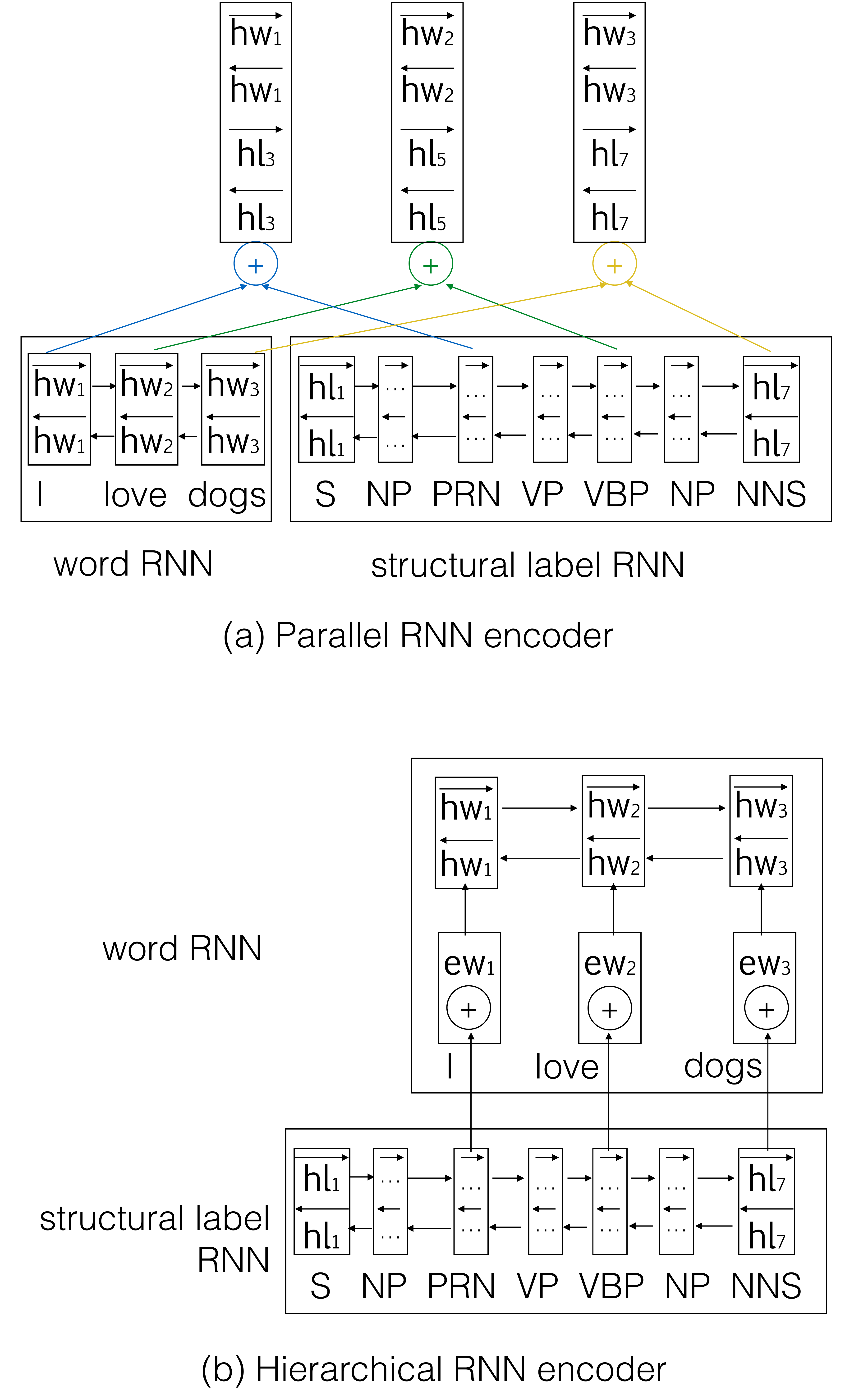}
\end{center}
\caption{The graphical illustration of the~\textit{Parallel RNN} encoder (a) and the \textit{Hierarchical RNN} encoder (b). Here, $\overrightarrow{hw_j}$ and $\overleftarrow{hw_j}$ are the forward and backward hidden states for word $w_j$, $\overrightarrow{hl_i}$ and $\overleftarrow{hl_i}$ are for structural label $l_i$, $ew_j$ is the word embedding for word $w_j$, and $\bigoplus$ is for concatenation operator.}\label{fig:two_rnn}
\end{figure}

There is no doubt that the structural label sequence is much longer than its word sequence. In order to obtain the structural label annotation vector for $w_i$ in word sequence, we simply look for $w_i$'s part-of-speech (POS) tag in the label sequence and view the tag's annotation vector as $w_i$'s label annotation vector. This is because $w_i$'s POS tag location can also represent $w_i$'s location in the parse tree. For example, in Figure~\ref{fig:parse_tree}, word $w_1$ in (a) maps to $l_3$ in (c) since $l_3$ is the POS tag of $w_1$. Likewise, $w_2$ maps to $l_5$ and $w_3$ to $l_7$. That is to say, we use $l_3$'s learned annotation vector as $w_1$'s label annotation vector. 

\subsection{RNN Encoders with Source Syntax}
In the next, we first propose two different encoders to augment word annotation vector with its corresponding label annotation vector, each of which consists of two RNNs~\footnote{Hereafter, we simplify bi-RNN as RNN.}: in one encoder, the two RNNs work independently (i.e., \textit{Parallel RNN Encoder}) while in another encoder the two RNNs work in a hierarchical way (i.e., \textit{Hierarchical RNN Encoder}). The difference between the two encoders lies in how the two RNNs interact. Then, we propose the third encoder with a single RNN, which learns word and label annotation vectors stitchingly (i.e., \textit{Mixed RNN Encoder}). Since any of the above three approaches focuses only on the encoder as to generate source annotation vectors along with structural information, we keep the rest part of the NMT models unchanged. 

\vspace{0.3cm}
\noindent\textbf{Parallel RNN Encoder} Figure~\ref{fig:two_rnn} (a) illustrates our \textit{Parallel RNN} encoder, which includes two parallel RNNs: i.e., a word RNN and a structural label RNN. On the one hand, the word RNN, as in conventional NMT models, takes a word sequence as input and output a word annotation vector for each word. On the other hand, the structural label RNN takes the structural label sequence of the word sequence as input and obtains a label annotation vector for each label. Besides, we concatenate each word's word annotation vector and its POS tag's label annotation vector as the final annotation vector for the word. For example, the final annotation vector for word \textit{love} in Figure~\ref{fig:two_rnn} (a) is $[\overrightarrow{hw_2}; \overleftarrow{hw_2};\overrightarrow{hl_5};\overleftarrow{hl_5}]$, where the first two subitems $[\overrightarrow{hw_2}; \overleftarrow{hw_2}]$ are the word annotation vector and the rest two subitems $[\overrightarrow{hl_5}; \overleftarrow{hl_5}]$ are its POS tag \textit{VBP}'s label annotation vector.\\

\vspace{0.3cm}
\noindent\textbf{Hierarchical RNN Encoder} Partially inspired by the model architecture of GNMT~\cite{wu_etal:16} which consists of multiple layers of LSTM RNNs, we propose a two-layer model architecture in which the lower layer is the structural label RNN while the upper layer is the word RNN, as shown in Figure~\ref{fig:two_rnn} (b). We put the word RNN in the upper layer because each item in the word sequence can map into an item in the structural label sequence, while this does not hold if the order of the two RNNs is reversed. As shown in Figure~\ref{fig:two_rnn} (b), for example, the POS tag \textit{VBP}'s label annotation vector ${[\overrightarrow{hl_5}, \overleftarrow{hl_5}]}$ is concatenated with word \textit{love}'s word embedding ${ew_2}$ to feed as the input to the word RNN. 

\begin{figure}
\begin{center}
\includegraphics[width=3.0in]{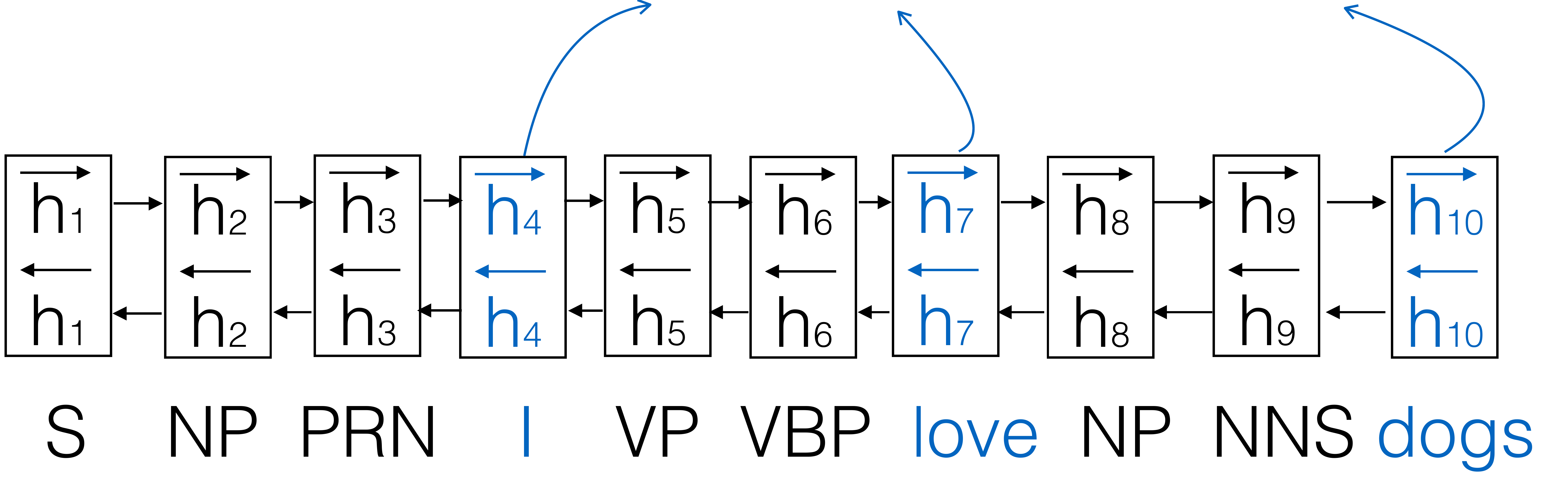}
\end{center}
\caption{The graphical illustration of the \textit{Mixed RNN} encoder. Here, $\overrightarrow{h_j}$ and $\overleftarrow{h_j}$ are the forward and backward hidden annotation vectors for \textit{j}-th item, which can be either a word or a structural label.}\label{fig:one_rnn}
\end{figure}

\vspace{0.3cm}
\noindent\textbf{Mixed RNN Encoder} Figure~\ref{fig:one_rnn} presents our \textit{Mixed RNN} encoder. Similarly, the sequence of input is the linearization of its parse tree (as in Figure~\ref{fig:parse_tree} (b)) following a depth-first traversal order, but being mixed with both words and structural labels in a stitching way. It shows that the RNN learns annotation vectors for both the words and the structural labels, though only the annotation vectors of words are further fed to decoding (e.g., $\big([\overrightarrow{h_4}, \overleftarrow{h_4}], [\overrightarrow{h_7}, \overleftarrow{h_7}], [\overrightarrow{h_{10}}, \overleftarrow{h_{10}}]\big)$). Even though the annotation vectors of structural labels are not directly fed forward for decoding, the error signal is back propagated along the word sequence and allows the annotation vectors of structural labels being updated accordingly. 

\subsection{Comparison of RNN Encoders with Source Syntax}
Though all the three encoders model both word sequence and structural label sequence, the differences lie in their respective model architecture with respect to the degree of coupling the two sequences:
\begin{itemize}
\item In the \textit{Parallel RNN} encoder, the word RNN and structural label RNN work in a parallel way. That is to say, the error signal back propagated from the word sequence would not affect the structural label RNN, and vice versa. In contrast, in the \textit{Hierarchical RNN} encoder, the error signal back propagated from the word sequence has a direct impact on the structural label annotation vectors, and thus on the structural label embeddings. Finally, the \textit{Mixed RNN} encoder ties the structural label sequence and word sequence together in the closest way. Therefore, the degrees of coupling the word and structural label sequences in these three encoders are like this: \textit{Mixed RNN} encoder \textgreater ~\textit{Hierarchical RNN} encoder \textgreater  ~\textit{Parallel RNN} encoder.

\item Figure~\ref{fig:two_rnn} and Figure~\ref{fig:one_rnn} suggest that the \textit{Mixed RNN} encoder is the simplest. Moreover, comparing to conventional NMT encoders, the difference lies only in the length of the input sequence. Statistics on our training data reveal that the \textit{Mixed RNN} encoder approximately triples the input sequence length compared to conventional NMT encoders.
\end{itemize} 

\section{Experimentation}\label{sect:exp}
We have presented our approaches to incorporating the source syntax into NMT encoders. In this section, we evaluate their effectiveness on Chinese-to-English translation.

\subsection{Experimental Settings}
Our training data for the translation task consists of 1.25M sentence pairs extracted from LDC corpora, with 27.9M Chinese words and 34.5M English words respectively.\footnote{The corpora include LDC2002E18, LDC2003E07, LDC2003E14, Hansards portion of LDC2004T07, LDC2004T08 and LDC2005T06.} We choose NIST MT 06 dataset (1664  sentence  pairs) as our development set, and NIST MT 02, 03, 04, and 05 datasets (878, 919, 1788 and 1082 sentence pairs, respectively) as our test sets.\footnote{\url{http://www.itl.nist.gov/iad/mig/tests/mt/}} To get the source syntax for sentences on the source-side, we parse the Chinese sentences with Berkeley Parser \footnote{\url{https://github.com/slavpetrov/berkeleyparser}}~\cite{petrov_klein:07} trained on Chinese TreeBank 7.0~\cite{xue_etal:05}. We use the case insensitive 4-gram NIST BLEU score~\cite{papineni_etal:02} for the translation task.

For efficient training of neural networks, we limit the maximum sentence length on both source and target sides to 50. We also limit both the source and target vocabularies to the most frequent 16K words in Chinese and English, covering approximately 95.8\% and 98.2\% of the two corpora respectively. All the out-of-vocabulary words are mapped to a special token \textit{UNK}. Besides, the word embedding dimension is 620 and the size of a hidden layer is 1000.  All the other settings are the same as in~\citeauthor{bahdanau_etal:15}\shortcite{bahdanau_etal:15}.

The inventory of structural labels includes 16 phrase labels and 32 POS tags. In both our \textit{Parallel RNN} encoder and \textit{Hierarchical RNN} encoder, we set the embedding dimension of these labels as 100 and the size of a hidden layer as 100. Besides, the maximum structural label sequence length is set to 100. In our \textit{Mixed RNN} encoder, since we only have one input sequence, we equally treat the structural labels and words (i.e., a structural label is also initialized with 620 dimension embedding). Compared to the baseline NMT model, the only different setting is that we increase the maximum sentence length on source-side from 50 to 150.  

We compare our method with two state-of-the-art models of SMT and NMT:
\begin{itemize}
\item cdec~\cite{dyer_etal:10}: an open source hierarchical phrase-based SMT system~\cite{chiang:07} with default configuration and a 4-gram language model trained on the target portion of the training data.\footnote{\url{https://github.com/redpony/cdec}}
\item RNNSearch: a re-implementation of the attentional NMT system ~\cite{bahdanau_etal:15} with slight changes taken from dl4mt tutorial.\footnote{\url{https://github.com/nyu-dl/dl4mt-tutorial}} For the activation function $f$ of an RNN, RNNSearch uses the gated recurrent unit (GRU) recently proposed by~\cite{cho_etal_emnlp:14}. It incorporates dropout~\cite{hinton_etal:12} on the output layer and improves the attention model by feeding the lastly generated word. We use AdaDelta~\cite{zeiler:12} to optimize model parameters in training with the mini-batch size of 80. For translation, a beam search with size 10 is employed. 
\end{itemize}

\subsection{Experiment Results}

\begin{table*}
\begin{center}

\begin{tabular}{r|r|r|r|l|lllll}
\hline
\textbf{\#} & \textbf{System} & \textbf{\#Params} & \textbf{Time} & \textbf{MT06} & \textbf{MT02} & \textbf{MT03} & \textbf{MT04} & \textbf{MT05} & \textbf{All} \\
\hline
1 & cdec & - & - & 33.4 & 34.8 & 33.0 & 35.7 & 32.1 & 34.2 \\
\hline
2 & RNNSearch & 60.6M & 153m & 34.0 & 36.9 & 33.7 & 37.0 & 34.1 & 35.6\\
\hline
3 & Parallel RNN & +1.1M & +9m &  34.8\dag & \textbf{37.8}\ddag & 34.2 & 38.3\ddag & 34.6 & 36.6\ddag\\
4 & Hierarchical RNN & +1.2M & +9m & 35.2\ddag & 37.2 & 34.7\dag & 38.7\ddag & 34.7\dag & 36.7\ddag\\
5 & Mixed RNN & +0 & +40m & \textbf{35.6}\ddag & 37.7\dag & \textbf{34.9}\ddag & \textbf{38.6}\ddag & \textbf{35.5}\ddag & \textbf{37.0}\ddag\\
\hline
\end{tabular}
\end{center}
\caption{\label{tbl:performance} Evaluation of the translation performance. \dag ~and \ddag: significant over RNNSearch at 0.05/0.01, tested by bootstrap resampling~\cite{koehn:04}.  “+” is the additional number of parameters or training time on the top of the baseline system RNNSearch. Column \textit{Time} indicates the training time in minutes per epoch for different NMT models}
\end{table*}

Table~\ref{tbl:performance} shows the translation performances measured in BLEU score. Clearly, all the proposed NMT models with source syntax improve the translation accuracy over all test sets, although there exist considerable differences among different variants. 

\vspace{0.2cm}
\noindent\textbf{Parameters} 
The three proposed models introduce new parameters in different ways. As a baseline model, RNNSearch has 60.6M parameters. Due to the infrastructure similarity, the \textit{Parallel RNN} system and the \textit{Hierarchical RNN} system introduce the similar size of additional parameters, resulting from the RNN model for structural label sequences (about 0.1M parameters) and catering either the augmented annotation vectors (as shown in Figure~\ref{fig:two_rnn} (a)) or the augmented word embeddings (as shown in Figure~\ref{fig:two_rnn} (b)) (the remain parameters). It is not surprising that the \textit{Mixed RNN} system does not require any additional parameters since though the input sequence becomes longer, we keep the vocabulary size unchanged, resulting in no additional parameters.

\vspace{0.2cm}
\noindent\textbf{Speed} Introducing the source syntax slightly slows down the training speed. When running on a single GPU GeForce GTX 1080, the baseline model speeds 153 minutes per epoch with 14K updates while the proposed structural label RNNs in both \textit{Parallel RNN} and \textit{Hierarchical RNN} systems only increases the training time by about 6\% (thanks to the small size of structural label embeddings and annotation vectors), and the \textit{Mixed RNN} system spends 26\% more training time to cater the triple sized input sequence.

\vspace{0.2cm}
\noindent\textbf{Comparison with the baseline NMT model (RNNSearch)} While all the three proposed NMT models outperform RNNSearch, the \textit{Parallel RNN} system and the \textit{Hierarchical RNN} system achieve similar accuracy (e.g., 36.6 \textit{v.s.} 36.7). Besides, the \textit{Mixed RNN} system achieves the best accuracy overall test sets with the only exception of NIST MT 02. Over all test sets, it outperforms RNNSearch by 1.4 BLEU points and outperforms the other two improved NMT models by 0.3$\sim$0.4 BLEU points, suggesting the benefits of high degree of coupling the word sequence and the structural label sequence. This is very encouraging since the \textit{Mixed RNN} encoder is the simplest, without introducing new parameters and with only slight additional training time. 

\vspace{0.2cm}
\noindent\textbf{Comparison with the SMT model (cdec)} Table~\ref{tbl:performance} also shows that all NMT systems outperform the SMT system. This is very consistent with other studies on Chinese-to-English translation~\cite{mi_etal:16,tu_etal:17,wang_etal:17}. 

\section{Analysis}\label{sect:dis}
As the proposed \textit{Mixed RNN} system achieves the best performance, we further look at the RNNSearch system and the \textit{Mixed RNN} system to explore more on how syntactic information helps in translation.

\subsection{Effects on Long Sentences}

\begin{figure}
\setlength{\abovecaptionskip}{0pt}
\begin{center}
\includegraphics[width=3.0in]{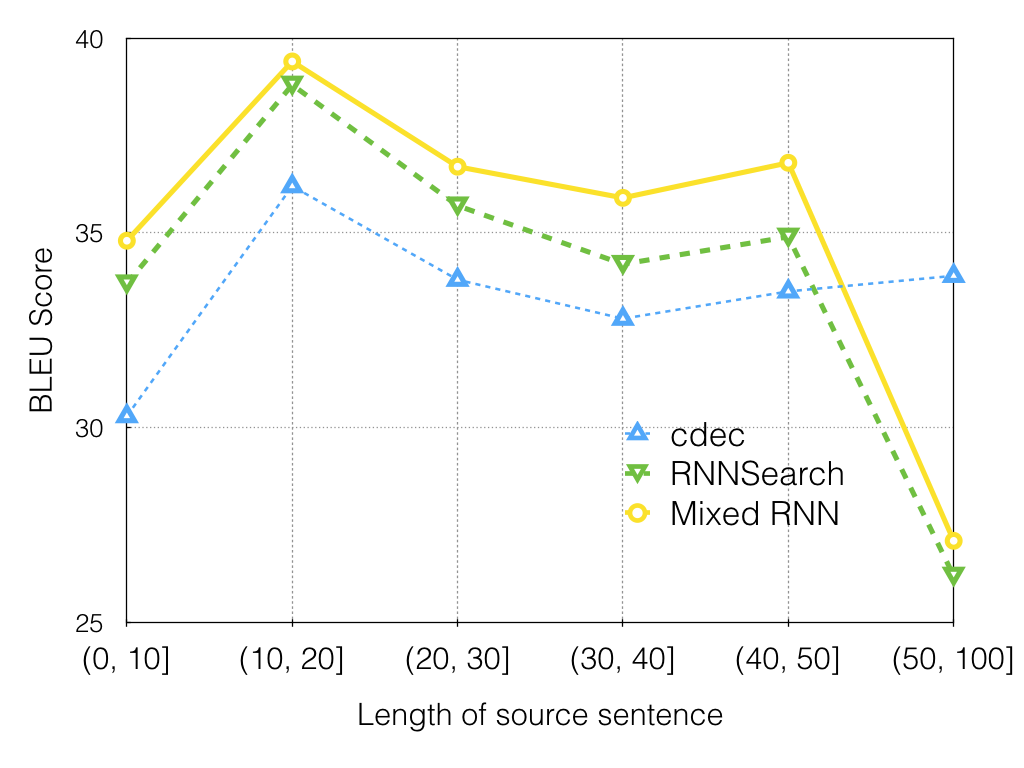}
\end{center}
\caption{Performance of the generated translations with respect to the lengths of the input sentences.}\label{fig:length}
\end{figure}

Following~\citeauthor{bahdanau_etal:15}~\shortcite{bahdanau_etal:15}, we group sentences of similar lengths together and compute BLEU scores. Figure~\ref{fig:length} presents the BLEU scores over different lengths of input sentences. It shows that \textit{Mixed RNN} system outperforms RNNSearch over sentences with all different lengths. It also shows that the performance drops substantially when the length of input sentences increases. This performance trend over the length is consistent with the findings in~\cite{cho_etal_sst:14,tu_etal:16,Tu:2017:TACL}.
We also observe that the NMT systems perform surprisingly bad on sentences over 50 in length, especially compared to the performance of SMT system (i.e., cdec). We think that the bad behavior of NMT systems towards long sentences (e.g., length of 50) is due to the following two reasons: (1) the maximum source sentence length limit is set as 50 in training,~\footnote{Though the maximum source length limit in \textit{Mixed RNN} system is set to 150, it approximately contains 50 words in maximum.} making the learned models not ready to translate sentences over the maximum length limit; (2) NMT systems tend to stop early for long input sentences. 

\subsection{Analysis on Word Alignment}
Due to the capability of carrying syntactic information in source annotation vectors, we conjecture that our model with source syntax is also beneficial for alignment. To test this hypothesis, we carry out experiments of the word alignment task on the evaluation dataset from~\citeauthor{liu_sun:15}~\shortcite{liu_sun:15}, which contains 900 manually aligned Chinese-English sentence pairs. We force the decoder to output reference translations, as to get automatic alignments between input sentences and their reference translations. To evaluate alignment performance, we report the alignment error rate (AER)~\cite{och_ney:03} in Table~\ref{tbl:aer}. 

\begin{table}
\begin{center}
\begin{tabular}{r|r}
\hline
\textbf{System} & \textbf{AER}\\
\hline
RNNSearch & 50.1 \\
\hline
Mixed RNN & 47.9 \\
\hline
\end{tabular}
\end{center}
\caption{\label{tbl:aer} Evaluation of alignment quality. The lower the score, the better the alignment quality.}
\end{table}

Table~\ref{tbl:aer} shows that source syntax information improves the attention model as expected by maintaining an annotation vector summarizing structural information on each source word. 

\subsection{Analysis on Phrase Alignment}
The above subsection examines the alignment performance at the word level. In this subsection, we turn to phrase alignment analysis by moving from word unit to phrase unit. Given a source phrase \textit{XP}, we use word alignments to examine if the phrase is translated continuously (\textbf{Cont.}), or discontinuously (\textbf{Dis.}), or if it is not translated at all (\textbf{Un.}). 

There are some phrases, such as noun phrases (NPs), prepositional phrases (PPs) that we usually expect to have a continuous translation. With respect to several such types of phrases, Table~\ref{tbl:phrase} shows how these phrases are translated. From the table, we see that translations of RNNSearch system do not respect source syntax very well. For example, in RNNSearch translations, 57.3\%, 33.6\%, and 9.1\% of PPs are translated continuously, discontinuously, and untranslated, respectively. Fortunately, our \textit{Mixed RNN} system is able to have more continuous translation for those phrases. Table~\ref{tbl:phrase} also suggests that there is still much room for NMT to show more respect to syntax. 

\begin{table}
\begin{center}

\begin{tabular}{r|r|rrr}
\hline
\textbf{System} & \textbf{XP} & \textbf{Cont.} & \textbf{Dis.} & \textbf{Un.}\\
\cline{2-5}
\hline
\multicolumn{1}{ r  }{\multirow{5}{*}{RNNSearch} } & \multicolumn{1}{ |r| }{PP} & 57.3 & 33.6 & 9.1 \\
\multicolumn{1}{ c  }{}&\multicolumn{1}{ |r| }{NP} & 59.8 & 25.5 & 14.7\\ 
\multicolumn{1}{ c  }{}&\multicolumn{1}{ |r| }{CP} & 47.3 & 44.6 & 8.1\\
\multicolumn{1}{ c  }{}&\multicolumn{1}{ |r| }{QP} & 54.0 & 22.2 & 23.8\\ 
\cline{2-5}
\multicolumn{1}{ c  }{}&\multicolumn{1}{ |r| }{ALL} & 58.1 & 27.1 & 14.8\\
\hline
\hline
\multicolumn{1}{ r  }{\multirow{5}{*}{Mixed RNN} } & \multicolumn{1}{ |r| }{PP} & 63.3 & 27.5 & 9.2 \\
\multicolumn{1}{ c  }{}&\multicolumn{1}{ |r| }{NP} & 63.1 & 23.1 & 13.8\\ 
\multicolumn{1}{ c  }{}&\multicolumn{1}{ |r| }{CP} & 54.5 & 36.6 & 8.9\\
\multicolumn{1}{ c  }{}&\multicolumn{1}{ |r| }{QP} & 56.2 & 19.7 & 24.1\\ 
\cline{2-5}
\multicolumn{1}{ c  }{}&\multicolumn{1}{ |r| }{ALL} & 60.4 & 25.0 & 14.6\\
\hline
\end{tabular}

\end{center}
\caption{\label{tbl:phrase} Percentages (\%) of syntactic phrases in our test sets being translated continuously, discontinuously, or not being translated. Here PP is for prepositional phrase, NP for noun phrase, CP for clause headed by a complementizer, QP for quainter phrase.}
\end{table}

\subsection{Analysis on Over Translation}
To estimate the over translation generated by NMT, we propose \textit{ratio of over translation (ROT)}:

\begin{small}
\begin{equation}
ROT = \frac{\sum_{w_i}{t(w_i)}}{|w|}
\end{equation}
\end{small}
\noindent where $|w|$ is the number of words in consideration, $t(w_i)$ is the times of over translation for word $w_i$. Given a word $w$ and its translation $e=e_1 e_2 \dots e_n$, we have:
\begin{small}
\begin{equation}
t(w) = |e| - |uniq(e)|
\end{equation}
\end{small}
\noindent where $|e|$ is the number of words in $w$'s translation $e$, while $|uniq(e)|$ is the number of unique words in $e$. For example, if a source word \textit{香港/xiangkang} is translated as \textit{hong kong hong kong}, we say it being over translated \textit{2} times.

\begin{table}
\begin{center}
\begin{tabular}{r|r|r}
\hline
\textbf{System} & \textbf{POS} & \textbf{ROT (\%)}\\
\cline{2-3}
\hline
\multicolumn{1}{ r  }{\multirow{5}{*}{RNNSearch} } & \multicolumn{1}{ |r| }{NR} & 15.7  \\
\multicolumn{1}{ c  }{}&\multicolumn{1}{ |r| }{CD} & 7.4 \\
\multicolumn{1}{ c  }{}&\multicolumn{1}{ |r| }{DT} & 4.9 \\
\multicolumn{1}{ c  }{}&\multicolumn{1}{ |r| }{NN} & 8.0 \\ 
\cline{2-3}
\multicolumn{1}{ c  }{}&\multicolumn{1}{ |r| }{ALL} & 5.5 \\
\hline
\hline
\multicolumn{1}{ r  }{\multirow{5}{*}{Mixed RNN} } & \multicolumn{1}{ |r| }{NR} & 12.3  \\
\multicolumn{1}{ c  }{}&\multicolumn{1}{ |r| }{CD} & 5.1 \\
\multicolumn{1}{ c  }{}&\multicolumn{1}{ |r| }{DT} & 2.4 \\
\multicolumn{1}{ c  }{}&\multicolumn{1}{ |r| }{NN} & 6.8 \\ 
\cline{2-3}
\multicolumn{1}{ c  }{}&\multicolumn{1}{ |r| }{ALL} & 4.5 \\
\hline
\end{tabular}

\end{center}
\caption{\label{tbl:over_translation} Ratio of over translation (ROT) on test sets. Here NR is for proper noun, CD for cardinal number, DT for determiner, and NN for nouns except proper nouns and temporal nouns.}
\end{table}

Table~\ref{tbl:over_translation} presents ROT grouped by some typical POS tags. It is not surprising that RNNSearch system has high ROT with respect to POS tags of NR (proper noun) and CD (cardinal number): this is due to the fact that the two POS tags include high percentage of unknown words which tend to be translated multiple times in translation. Words of DT (determiner) are another source of over translation since they are usually translated to multiple \textit{the} in English. It also shows that by introducing source syntax, \textit{Mixed RNN} system alleviates the over translation issue by 18\%: ROT drops from 5.5\% to 4.5\%.

\subsection{Analysis on Rare Word Translation}
We analyze the translation of source-side rare words that are mapped to a special token \textit{UNK}. Given a rare word \textit{w}, we examine if it is translated into a \textit{non-UNK} word (\textbf{non-UNK}), \textit{UNK} (\textbf{UNK}), or if it is not translated at all (\textbf{Un.}). 

\begin{table}
\begin{center}

\begin{tabular}{r|r|rrr}
\hline
\textbf{System} & \textbf{POS} & \textbf{non-UNK} & \textbf{UNK} & \textbf{Un.}\\
\cline{2-5}
\hline
\multicolumn{1}{ r }{\multirow{5}{*}{RNNSearch} } & \multicolumn{1}{ |r| }{NN} & 27.2 & 40.4 & 32.4 \\
\multicolumn{1}{ c }{}&\multicolumn{1}{ |r| }{NR} & 22.9 & 58.5 & 18.6\\ 
\multicolumn{1}{ c }{}&\multicolumn{1}{ |r| }{VV} & 34.5 & 32.9 & 32.7\\
\multicolumn{1}{ c }{}&\multicolumn{1}{ |r| }{CD} & 10.7 & 63.4 & 25.9\\ 
\cline{2-5}
\multicolumn{1}{ c }{}&\multicolumn{1}{ |r| }{ALL} & 27.2 & 40.4 & 32.4\\
\hline
\hline
\multicolumn{1}{ r }{\multirow{5}{*}{Mixed RNN} } & \multicolumn{1}{ |r| }{NN} & 24.8 & 41.4 & 33.8 \\
\multicolumn{1}{ c }{}&\multicolumn{1}{ |r| }{NR} & 17.0 & 64.5 & 18.6\\ 
\multicolumn{1}{ c }{}&\multicolumn{1}{ |r| }{VV} & 33.6 & 34.0 & 32.3\\
\multicolumn{1}{ c }{}&\multicolumn{1}{ |r| }{CD} & 9.6 & 68.7 & 21.7\\ 
\cline{2-5}
\multicolumn{1}{ c }{}&\multicolumn{1}{ |r| }{ALL} & 23.9 & 47.5 & 28.7\\
\hline
\end{tabular}

\end{center}
\caption{\label{tbl:rare_word} Percentages (\%) of rare words in our test sets being translated into a \textit{non-UNK} word (\textbf{non-UNK}), \textit{UNK} (\textbf{UNK}), or if it is not translated at all (\textbf{Un.}).}
\end{table}

Table~\ref{tbl:rare_word} shows how source-side rare words are translated. The four POS tags listed in the table account for about 90\% of all rare words in the test sets. It shows that in \textit{Mixed RNN} system is more likely to translate source-side rare words into \textit{UNK} on the target side. This is reasonable since the source side rare words tends to be translated into rare words in the target side. Moreover, it is hard to obtain its correct non-\textit{UNK} translation when a source-side rare word is replaced as \textit{UNK}. 

Note that our approach is compatible with with approaches of open vocabulary. Taking the subword approach~\citep{sennrich_etal:16} as an example, for a word on the source side which is divided into several subword units, we can synthesize sub-POS nodes that cover these units. For example, if \textit{misunderstand/VB} is divided into units of \textit{mis} and \textit{understand}, we construct substructure \textit{(VB (VB-F mis) (VB-I understand))}.

\section{Related Work}\label{sect:related}
While there has been substantial work on linguistically motivated SMT, approaches that leverage syntax for NMT start to shed light very recently. Generally speaking, NMT can provide a flexible mechanism for adding linguistic knowledge, thanks to its strong capability of automatically learning feature representations.

\citeauthor{eriguchi_etal:16}~\shortcite{eriguchi_etal:16} propose a tree-to-sequence model that learns annotation vectors not only for terminal words, but also for non-terminal nodes. They also allow the attention model to align target words to non-terminal nodes. Our approach is similar to theirs by using source-side phrase parse tree. However, our \textit{Mixed RNN} system, for example, incorporates syntax information by learning annotation vectors of syntactic labels and words stitchingly, but is still a sequence-to-sequence model, with no extra parameters and with less increased training time. 

\citeauthor{sennrich_haddow:16}~\shortcite{sennrich_haddow:16} define a few linguistically motivated features that are attached to each individual words. Their features include lemmas, subword tags, POS tags, dependency labels, etc. They concatenate feature embeddings with word embeddings and feed the concatenated embeddings into the NMT encoder. On the contrast, we do not specify any feature, but let the model implicitly learn useful information from the structural label sequence. 

\citeauthor{shi_etal:16}~\shortcite{shi_etal:16} design a few experiments to investigate if the NMT system without external linguistic input is capable of  learning syntactic information on the source-side as a by-product of training. However, their work is not focusing on improving NMT with linguistic input. Moreover, we analyze what syntax is disrespected in translation from several new perspectives.

\citeauthor{garcia_martinez_etal:16}~\shortcite{garcia_martinez_etal:16} generalize NMT outputs as lemmas and morphological factors in order to alleviate the issues of large vocabulary and out-of-vocabulary word translation. The lemmas and corresponding factors are then used to generate final words in target language. Though they use linguistic input on the target side, they are limited to the word level features. Phrase level, or even sentence level linguistic features are harder to obtain for a generation task such as machine translation, since this would require incremental parsing of the hypotheses at test time.

\section{Conclusion}\label{sect:con}
In this paper, we have investigated whether and how source syntax can explicitly help NMT to improve its translation accuracy. 

To obtain syntactic knowledge, we linearize a parse tree into a structural label sequence and let the model automatically learn useful information through it. Specifically, we have described three different models to capture the syntax knowledge, i.e., \textit{Parallel RNN}, \textit{Hierarchical RNN}, and \textit{Mixed RNN}. Experimentation on Chinese-to-English translation shows that all proposed models yield improvements over a state-of-the-art baseline NMT system. It is also interesting to note that the simplest model (i.e., \textit{Mixed RNN}) achieves the best performance, resulting in obtaining significant improvements of 1.4 BLEU points on NIST MT 02 to 05. 

In this paper, we have also analyzed the translation behavior of our improved system against the state-of-the-art NMT baseline system from several perspectives. Our analysis shows that there is still much room for NMT translation to be consistent with source syntax. In our future work, we expect several developments that will shed more light on utilizing source syntax, e.g., designing novel syntactic features (e.g., features showing the syntactic role that a word is playing) for NMT, and employing the source syntax to constrain and guild the attention models. 

\section*{Acknowledgments}
The authors would like to thank three anonymous reviewers for providing helpful comments, and also acknowledge Xing Wang, Xiangyu Duan, Zhengxian Gong  for useful discussions. This work was supported by National Natural Science Foundation of China (Grant No. 61525205, 61331011, 61401295). 

\balance
\bibliographystyle{acl2017}
\bibliography{nmt}

\end{CJK}
\end{document}